\documentclass[a4paper]{article}
\usepackage{mathtools,amsmath}
\usepackage{INTERSPEECH2019}
\usepackage{url}
\title{Learning to Recognise Words using Visually Grounded Speech}
\name{Sebastiaan Scholten$^1$, Danny Merkx$^2$, Odette Scharenborg$^1$}
\address{
  $^1$Multimedia Computing Group, Delft University of Technology, Delft, the Netherlands\\
  $^2$Centre for Language Studies, Radboud University, Nijmegen, the Netherlands}
\email{J.S.M.Scholten@student.tudelft.nl, D.Merkx@let.ru.nl, O.E.Scharenborg@tudelft.nl}

\begin{document}

\maketitle
\begin{abstract}
We investigated word recognition in a Visually Grounded Speech model. The model has been trained on pairs of images and spoken captions to create visually grounded embeddings which can be used for speech to image retrieval and vice versa. We investigate whether such a model can be used to recognise words by embedding isolated words and using them to retrieve images of their visual referents. We investigate the time-course of word recognition using a gating paradigm and perform a statistical analysis to see whether well known word competition effects in human speech processing influence word recognition. Our experiments show that the model is able to recognise words, and the gating paradigm reveals that words can be recognised from partial input as well and that recognition is negatively influenced by word competition from the word initial cohort.

\end{abstract}

\section{Introduction}
Babies initially have little semantic understanding of what is being said around them. It is theorized that the fact that they repeatedly hear certain words while they observe certain objects around them enables them to learn a mapping between speech and objects \cite{rasanen2015joint}.
Repetitive hearing of these utterances in the context of some functional consistency, such as picking up an object, will display the meaning of a smaller constituent of such an utterance, e.g., a word, and potentially about the class of objects it belongs to \cite{tomasello2009usage}.

Some core principles of Visually Grounded Speech (VGS) models are inspired by such learning processes. While most speech recognition research focuses on speech signals only, Visually Grounded Speech models include visual information rather than textual transcriptions to guide the training of the acoustic models \cite{harwath2015deep,harwath2016unsupervised,chrupala2017representations,merkx2019language,kamper2017semantic,scharenborg2020,kamper2018visually}
. Following the approach of multimodal neural models which produce visual-semantic alignments for images and text \cite{karpathy2015deep}, a VGS model employs two parallel Deep Neural Networks (DNNs) which are trained to map a speech signal and a corresponding image into a common embedding space.

Recent research on VGS models has seen an improvement in architectures and training schemes  \cite{chrupala2017representations,harwath2018jointly,merkx2019language} and different applications of the VGS model have been proposed such as semantic keyword spotting \cite{kamper2017semantic,kamper2019semantic}
and speech-based image retrieval \cite{harwath2015deep,chrupala2017representations,merkx2019language,scharenborg2020}, where a trained VGS model is fed full sentence speech captions with which the model retrieves the corresponding image. Recent research has shown that VGS models implicitly learn to recognise meaningful sentence constituents such as phonemes and words and the presence of these constituents can be decoded from the speech embeddings \cite{chrupala2017representations,merkx2019language,havard2019word,harwath2019learning, chrupala2020analyzing}. Havard and colleagues presented isolated words to a VGS model and investigated whether the model was able to retrieve images of the words' correct visual referents \cite{havard2019word}. This showed that the model does not just encode these constituents into the speech embeddings, but the model actually `recognises' individual words and learned to map them onto their correct visual referents.

Building on the synthetic speech experiments by Havard and colleagues, we investigate how natural speech is recognised by a VGS model using real human speech. In this paper, we will 1) investigate isolated word recognition using real speech, 2) investigate how words are recognised by a VGS model over time, 3) and look more in depth into the linguistic and acoustic properties that aid or hinder word recognition. As in \cite{havard2019word}, we use the retrieval of images containing a word's correct visual referent as a measure of the model's word recognition performance. Word recognition is expected to be more challenging with real speech as opposed to synthetic speech, due to real speech having more variation in quality, noise and speaking rate  synthetic speech. This can also be seen in \cite{chrupala2017representations}, where the model trained on real speech performs significantly worse with caption-to-image retrieval than a model trained on synthetic speech.

We carry out two experiments, inspired by those of \cite{havard2019word}. In our first experiment, the VGS model is fed individual words, which will allow us to investigate whether the model is actually learning to recognise individual words, which would be shown by the model being able to retrieve a relevant image on the basis of a single word rather than the full caption. In the second experiment, we use a gating paradigm, borrowed from human speech processing research. In the gating experiment, a word is presented to the VGS model in speech segments of increasing
duration, i.e., with increasing number of phones, and `asked' to retrieve an image of the correct visual referent on the basis of the available phone string. This allows us to investigate 1) the time-course of word recognition, 2) the amount of information needed for word recognition, and 3) whether the model is able to encode phones in the combined embedding space.

To answer our third question, we carry out a statistical analysis in which word recognition performance is predicted using several linguistic and acoustic features. These linguistic and acoustic features are factors known to influence human speech processing. In human speech processing (see for an overview of models of human speech processing Weber \& Scharenborg \cite{weber2012models}), the incoming speech signal is mapped against phone representations in the listener's brain, and the sounds that best resemble the incoming speech signal are `activated'. These activated phone representations,
activate every possible word in which they appear, irrespective of the position of the phone in the word. 
As more speech information becomes available, words that no longer match the input will drop out of the list of activated words. The word that best matches the speech input is recognised. Words that are activated are called competitors or competitor words. 
The number of competitor words plays a role in human speech processing: the more competitors there are, the longer it takes for a word to be recognised \cite{norris1995competition}. We want to see whether our VGS model activates competitor words in a similar manner, which would be shown by a significant effect of the number of competitor words on word recognition performance. We focus on the number of words that share the start of the word, the so-called word-initial cohort, as we are testing isolated words in our experiment \cite{marslen1978processing}, and the neighbourhood density, i.e., the number of words that differs exactly one phoneme from the target word.

The rest of this paper is organised as follows. Firstly, we discuss the model architecture and the methodology behind the experiments. Secondly, the results for the different experiments will be discussed. Lastly, this work will be concluded with a discussion with a summary of the contributions, as well as recommendations for future research.

\section{Methodology}

\subsection{Visually Grounded Speech Model}

\begin{figure}[t]
  \centering
  \includegraphics[width=\linewidth]{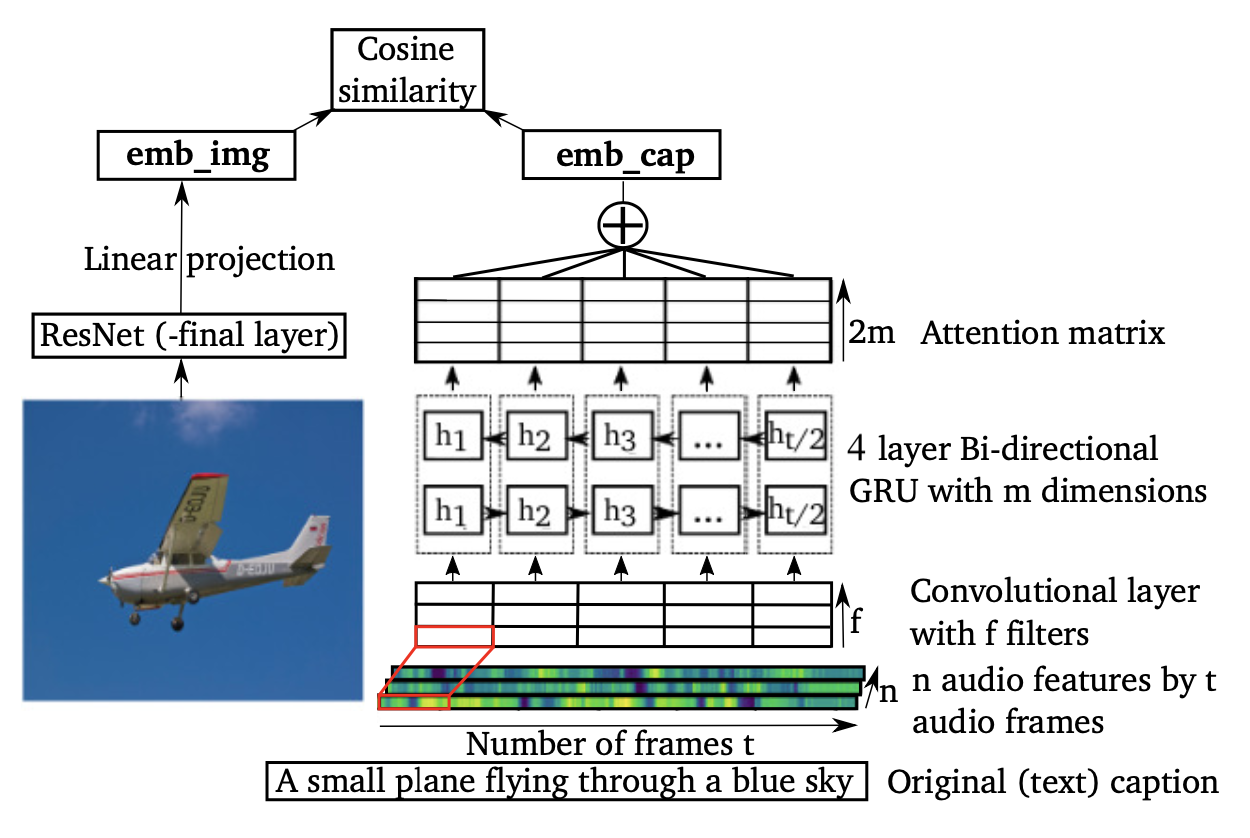}
  \caption{A visual representation of the image encoder parallel to the caption encoder. Based on \cite{merkx2019language}.}
  \label{fig:representation}
\end{figure}

For this paper, we use the Visually Grounded Speech Model implementation presented in \cite{merkx2019language}, with the addition of an extra Gated Recurrent Unit (GRU) layer, which can improve the the model's ability to capture long-range dependencies. The model consists of two DNNs: a pretrained image encoder and a Recurrent Neural Network (RNN)-based speech caption encoder. The encoders embed the speech and images, and the model is trained to minimise the cosine distance between image-caption pairs in the shared embedding space. A visual representation of the model is given in Figure \ref{fig:representation}.

The pre-trained image encoder is ResNet-152, which was trained on ImageNet \cite{he2016deep}.
The final object classification layer is removed from this network, and we place a single linear layer on top of ResNet and L2 normalise the result to map the latent image features onto our multimodal embedding space.

Our audio features consist of Mel Frequency Cepstral Coefficients (MFCCs). A 39-dimensional feature vector was used, comprising of 12 MFCCs including their log energy feature and first and second derivatives. A 1-dimensional convolutional layer was applied to the 39-dimensional feature vector, then these channels were fed to an RNN with a 4-layer bi-directional GRU. Then, the 1024 bi-directional units were concatenated to create a 2048 feature vector, which feeds into a self-attention layer. The resulting feature representations are L2 normalised to arrive at the final caption embedding.

The caption encoder was trained in order for the image and speech pairs to have a cosine similarity larger by a margin $\alpha$ than the cosine similarity for mismatched pairs. We used a hinge loss function to minimise cosine distance for ground-truth pairs. The model was trained for 32 epochs with a batch size of 32. For a more detailed description of the model and the loss function please refer to  \cite{merkx2019language}.

We train the model on Flickr8k \cite{hodosh2013framing}, a database with 8k images and 5 written captions per image for a total of 40k captions. Harwath and colleagues collected spoken versions of these captions from a total of 183 different speakers, with a vocabulary of 8918 unique words \cite{harwath2015deep}. For our training, validation and test set we make use of the data split provided by \cite{karpathy2015deep}. We use spoken caption-to-image retrieval to evaluate how well our model performs on the training task and compare the model with previous work. Caption-to-image retrieval is measured in Recall@N, the percentage of captions for which the correct image was in the top N retrieved images. Images are retrieved based on their embedding distance to the caption embedding.

\subsection{Experiments}
We will be performing two experiments. In the first experiment, we present our model with isolated words to investigate how well the model learned to map these words onto their visual referents. In the second experiment, we segment the words into phonemes and present our model with phoneme sequences of increasing length to investigate the time-course of word recognition in the model. We present our model with multiple instances of each word, spoken by different speakers to gain a more realistic impression of how a word performs across different speakers and contexts. Also, this allows us to test which acoustic factors in the speech signal are influencing the model's word recognition performance.

\subsubsection{Experimental data}

A visually grounded model relies on there being a consistency between the image and speech signal in order to create a common embedding space. Therefore, we chose 49 words with clear visual referents, such as `bike' and `man', as opposed to articles and adverbs.  We extracted 50 occurrences of each word from the speech captions in the test set, to have an equal sample size for each word to allow a fair comparison between their word recognition performance.

The words were extracted from the speech signal using a forced alignment of the phonetic transcriptions with the speech captions in Flick8k. For the second experiment, these words were segmented into sequences of phonemes where each sequence was one phoneme longer than the previous. For example, for the word `bike', the speech signal was segmented into `B', `B-AY', and `B-AY-K'.

\subsubsection{Evaluating word recognition performance}

Following \cite{havard2019word}, we use the retrieval of images containing a word's correct visual referent as a measure of the model's word recognition performance. In order to quantify this we use the Precision@10 score which is calculated as follows. We use the trained VGS model to create embeddings for all of the word instances. From the Flickr8k test set, we take all images which had one of our 49 words in its captions and use the VGS model to create image embeddings. For each embedded word instance we then retrieve the ten most similar image embeddings as defined by cosine similarity between the embeddings.
The Precision@10 (P@10) is then calculated for each word instance as the percentage of its top ten images which contain the correct visual referent of the word.

\subsubsection{Evaluating linguistic and acoustic factors}\label{lmer}

To answer our third research question, we examine linguistic and acoustic factors which might influence the model's word recognition performance using a Linear Mixed Effects Regression (LMER). For the LMER analysis we used the \textit{lme4} package in R \cite{lme4}. All fixed effects are z-score normalised. The dependent variable is the P@10 score.

For the word recognition experiment, our LMER model takes into consideration the signal duration (i.e., number of speech frames), the speaking rate calculated as the number of phonemes in the word divided by its signal duration, the frequency of occurrence of the word in the training set and the number of phonemes, vowels, and consonants in the word. We also included the two-way interaction of the frequency of occurrence of the word in the training set with the number of phonemes, vowels, and consonants. We considered these interaction effects because words with a certain number of phonemes, vowels, and consonants might appear more often in a dataset. Furthermore, we included by-speaker and by-word random intercepts and by-speaker random slopes for the signal length, to take into consideration speaker differences on the duration of the signal.

For the second experiment, the LMER model takes into account the earlier mentioned frequency of occurrence of the word in the training set and the total number of phonemes in the word. We also include the size of the word-initial cohort and neighbourhood density. The word-initial cohort is calculated by determining for each phoneme sequence the number of words which start with the same phoneme sequence in the Flickr8k training set, which considers a total of 6182 unique words. This indicates the number of words that is considered simultaneously for recognition by the model given the phoneme sequence seen so far. The neighbourhood density is calculated as the number of words from the words in the Flickr8k training set that can be formed from the phoneme sequence by a one-phoneme substitution \cite{vitevitch2016phonological}. This factor indicates the similarity among spoken forms of words, and is therefore a second measure of the number of words that are simultaneously considered for recognition. The model also includes a by-speaker and a by-word random intercept.

\begin{table}[b]
  \caption{Speech caption-to-image retrieval scores including 95\% confidence intervals for our model. For comparison, the models of Merkx et al. \cite{merkx2019language}, Chrupa{\l}a et al. \cite{chrupala2017representations} and Harwath et al. \cite{harwath2015deep} which were also trained on Flickr8k speech captions are provided.}
  \label{tab:example}
  \centering
  \begin{tabular}{l r r r r}
    \toprule
    \multicolumn{1}{l}{\textbf{Model}} & \multicolumn{1}{c}{
    R@1} & \multicolumn{1}{c}{R@5}& \multicolumn{1}{c}{R@10}& \multicolumn{1}{c}{Med. R} \\
    \midrule
4-GRU & 10.71$\pm$1.9      & 29.2$\pm$2.8      & 40.2$\pm$3.0      & 18 \\
\cite{merkx2019language}& 8.0$\pm$1.7      & 24.5$\pm$2.7      & 35.5$\pm$3.0      & 24 \\
\cite{chrupala2017representations} & 5.5$\pm$1.4     & 16.3$\pm$2.3      & 25.3$\pm$2.7      & 48 \\
\cite{harwath2015deep} & \- & \- & 17.9$\pm$2.4 & \- \\
    \bottomrule
  \end{tabular}
\label{fig:modelscores}
\end{table}
\section{Results}

The scores in Table \ref{fig:modelscores} show the result for the speech caption-to-image retrieval task. This indicates how well the model learned to embed the speech and images in the common embedding space. R@N is the percentage of items for which the correct image was in the top N retrievals. Median R is the median rank of the correctly retrieved image. The addition of an extra GRU layer has led to a substantial performance increase, allowing dependencies in longer speech captions to be captured better.

\subsection{Word recognition}

\begin{figure}[t]
  \centering
  \includegraphics[width=\linewidth]{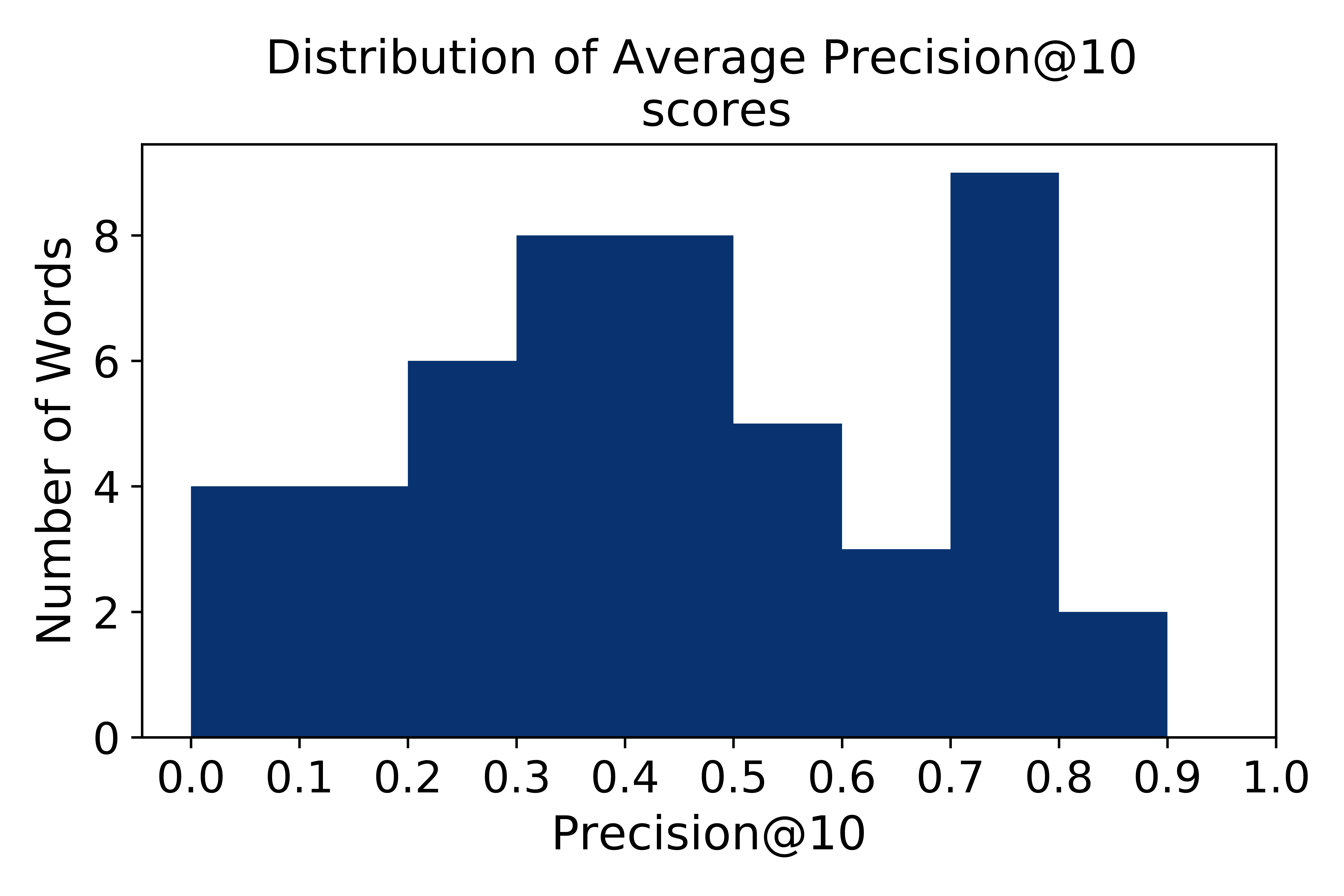}
  \caption{Distribution of Average P@10 scores for the 49 tested words, assigned to bin intervals of size 0.1.}
  \label{fig:precision}
\end{figure}

In this experiment, we present isolated words to the model. The histogram in Figure \ref{fig:precision} shows the distribution of the P@10 scores over the 49 words. The average P@10 is 0.44, which indicates that on average 4.4 out of the ten retrieved images contain the correct visual referent. However, Figure \ref{fig:precision} also shows that four words have a P@10 near zero, meaning that no correct images were retrieved and the word was not recognised. Furthermore, Havard and colleagues \cite{havard2019word} reported a median P@10 of 0.8, while we on the other hand have a median P@10 of 0.4. While our model does learn to recognise most words to some degree, this indicates a large difference in recognition performance going from the synthetic speech dataset in \cite{havard2019word} to the real speech of Flickr8k.

\begin{table}[b]
  \caption{Significant fixed effects with Standard Errors for the word recognition LMER.}
  \label{tab:example}
  \centering
  \begin{tabular}{l r r}
    \toprule
    \multicolumn{1}{l}{\textbf{Fixed effects}} & \multicolumn{1}{c}{
    Estimate} & \multicolumn{1}{c}{P-value}\\
    \midrule
        Intercept & 0.432$\pm$0.033 & $\textless$0.001 \\
        Signal duration & -0.050$\pm$0.014 & $\textless$0.001 \\
        Speaking rate & -0.068$\pm$0.013 & $\textless$0.001 \\
        Training set frequency & 0.152$\pm$0.063 & 0.020 \\
    \bottomrule
  \end{tabular}
\label{fig:experiment1}
\end{table}

Table \ref{fig:experiment1} shows the results from the statistical test. Firstly, signal duration was found to have a significant negative effect on the P@10 scores. This shows that the model has more difficulty encoding longer words. Secondly, speaking rate also had a significant negative effect, showing that words that are spoken more rapidly were encoded less well than words pronounced more slowly. Lastly, the frequency of occurrence of the word in the training set was shown to have a significant positive effect on word recognition performance. This shows that words which occur more often in training samples are encoded considerably better for word recognition. No interaction effects were found.

For our random effects, we see that the standard deviation of the scores between words is far larger than between speakers. This shows that the effect of using different speakers causes less variation in results in comparison to using different words.

\subsection{Word activation}

\begin{figure}[t]
  \centering
  \includegraphics[width=\linewidth]{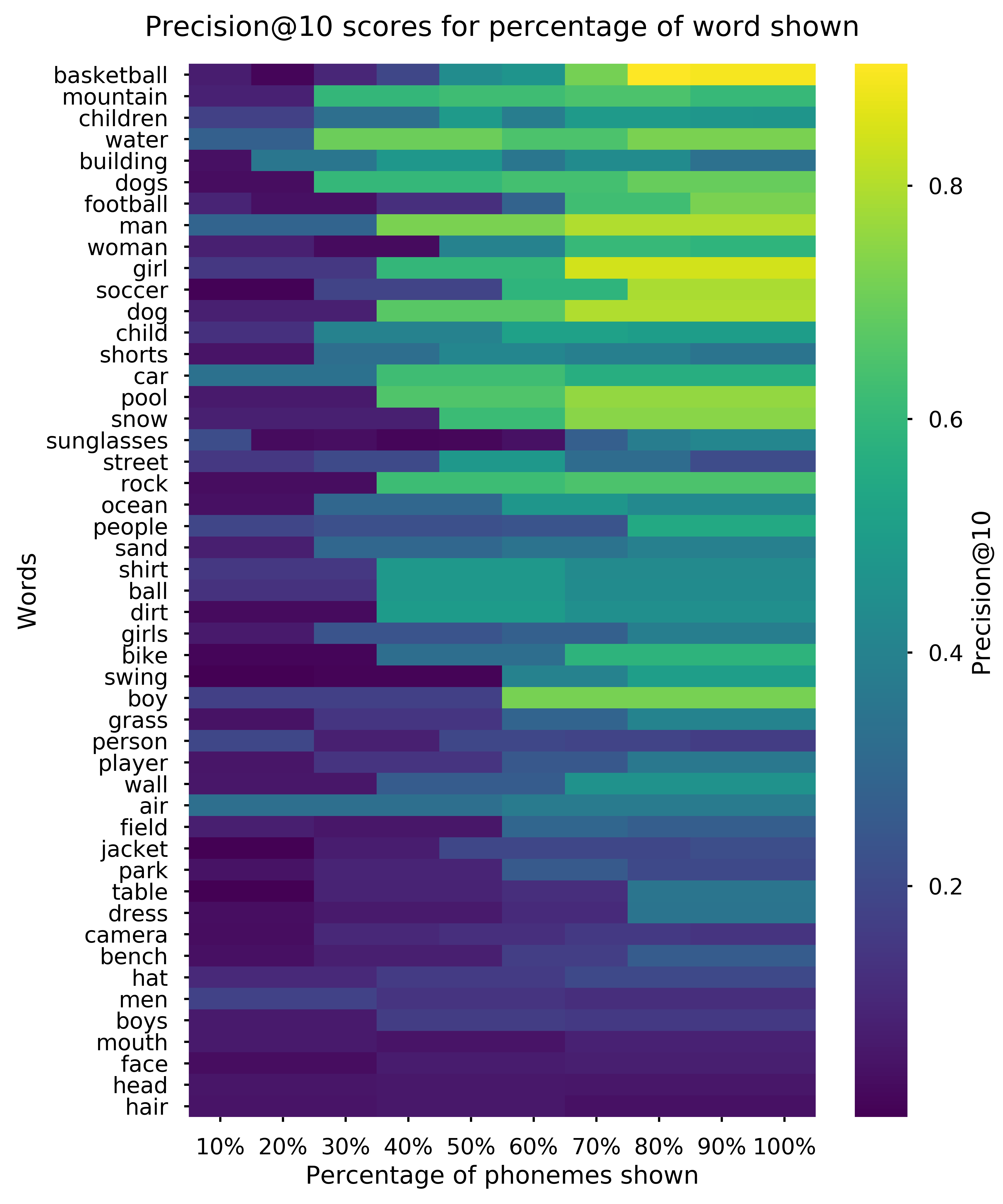}
  \caption{Heatmap showing the P@10 scores of a given word (shown on the y-axis) as a function of the phoneme sequence length. The x-axis indicates the percentage of phonemes of the word that were available to the model.}
  \label{fig:heatmap}
\end{figure}

In order to investigate the time-course of word recognition and how much information is needed for word recognition, phoneme sequences of increasing length were given to the model. Figure \ref{fig:heatmap} shows the results in terms of the P@10 of a given word (shown on the y-axis) as a function of the phoneme sequence length in terms of percentage of phonemes of the word. Note that the x-axis has ten values, if a word has for instance only two phonemes, the P@10 for the first and second phoneme span 10-50\% and 60-100\% respectively. A more yellow colour corresponds to a higher P@10.

As can be seen in Figure \ref{fig:heatmap}, generally, the more phonemes of a word the model is exposed to, the better it can retrieve the image corresponding to the spoken word. Some words representations, see the bottom of Figure 3 (bars are entirely blue), are not recognised at all irrespective of the percentage of phonemes shown to the model.

The results of the LMER model are summarised in Table \ref{fig:experiment2}. Unsurprisingly, the number of phonemes in a phoneme sequence has a significant positive effect on the P@10 scores indicating that words are recognised better when the model is presented with longer phoneme sequences. The frequency of occurrence of the word in the training set again has a significant positive effect on the performance, showing that having more training examples allows phoneme sequences to be mapped more easily to the correct visual referent. The word-initial cohort has a significant negative effect on the P@10 scores, indicating that, similar to human listeners, word recognition is more difficult when there are more words that have the same phoneme sequence at the start of the word. The effect of the neighbourhood density was not found to be significant.

\begin{table}[t]
  \caption{Significant fixed effects with Standard Errors for the word activation LMER.}
  \label{tab:example}
  \centering
  \begin{tabular}{l r r}
    \toprule
    \multicolumn{1}{l}{\textbf{Fixed effects}} & \multicolumn{1}{c}{
    Estimate} & \multicolumn{1}{c}{P-value}\\
    \midrule
        Intercept & 0.295$\pm$0.020&$\textless$0.001\\
        \# of phonemes & 0.134$\pm$0.003 & $\textless$0.001 \\
        Training set frequency & 0.087$\pm$0.018 & $\textless$0.001 \\
        Word-initial cohort  & -0.037$\pm$0.003 & $\textless$0.001 \\
    \bottomrule
  \end{tabular}
\label{fig:experiment2}
\end{table}

\section{Discussion and Conclusions}

In this paper, we investigated how natural speech is recognised by a Visually Grounded Speech model using real human speech. In order to do this, in the first experiment, we investigated how isolated words are recognized in a VGS model. Although our model is trained on full speech captions, the word recognition experiment showed that the model learned to recognise individual words and was able to map them onto their correct visual referent in most cases.

Also, we investigated the time course of the word recognition. The second experiment showed that it is possible to recognise a word from only a partial phoneme sequence and that word recognition performance (as measured in image retrieval scores) generally improved as more phonemes were seen, with the best retrieval scores when the model was shown all phonemes of the word. The largest leap in word recognition performance was observed after the model was provided with a phoneme sequence consisting of 30\%-40\% of the target word's phonemes. For some words such as `person' or `men', word recognition was highest right after the first phoneme and decreased upon seeing more of the speech signal, although in these cases the word generally was not recognised well. Similar to human listeners \cite{weber2012models}, the model did not need to have available all phonemes of the word in order to recognize it, which indicates that the model encodes useful information at the phoneme level.

Lastly, we looked in more depth at which linguistic and acoustic features influence word recognition performance. In general, words that are spoken more slowly have a higher word recognition score. The effect of frequency of a word in the training set on word recognition performance demonstrates how reliant such a model is on its training data. Furthermore, the size of the word-initial cohort was found to have a significant effect on word recognition performance. This shows that, similar to human speech processing, the number of words that match the input speech influence recognition accuracy. It is well known that in human speech recognition, words can be activated or suppressed by priming effects, thus hindering or aiding in recognition \cite{chiappe1996semantic}. It would be an interesting direction for future research to see if words preceded by a priming context show the expected effects on word recognition performance.

For future research it would be interesting to look at what word a sequence of phonemes is mapped to when it does not retrieve the correct image. This could give more insight into how phonemes are embedded within the model. Also, it would be interesting to see if there are other linguistic or acoustic factors in addition to those we investigated which affect word recognition performance.

\newpage
\bibliographystyle{IEEEtran}

\bibliography{mybib}

\end{document}